# TAPE: Tether-Aware Path Planning for Autonomous Exploration of Unknown 3D Cavities using a Tangle-compatible Tethered Aerial Robot

Louis Petit[1] and Alexis Lussier Desbiens[1]

*Abstract*—**This paper presents the first method for autonomous exploration of unknown cavities in three dimensions (3D) that focuses on minimizing the distance traveled and the length of tether unwound. Considering that the tether entanglements are little influenced by the global path, our approach employs a 2-level hierarchical architecture. The global frontier-based planning solves a Traveling Salesman Problem (TSP) to minimize the distance. The local planning attempts to minimize the path cost and the tether length using an adjustable decision function whose parameters play on the trade-off between these two values. The proposed method, TAPE, is evaluated through detailed simulation studies as well as field tests. On average, our method generates a 4.1% increase in distance traveled compared to the TSP solution without our local planner, with which the length of the tether remains below the maximum allowed value in 53% of the simulated cases against 100% with our method.**

*Index Terms*—**Motion and Path Planning, Aerial Systems: Perception and Autonomy.**

## I. INTRODUCTION

AUTONOMOUS robotic exploration is a major field of research and has led to many uses in areas such as search and rescue [1], inspection [2] and surveillance [3]. The main objective of the exploration of a priori unknown cavities is to cover the environment completely, in the most efficient way possible. To do so, a robot is equipped with sensors, such as cameras or LiDARs. Indoor environments are often GPS-denied. The data obtained from the sensors is therefore used to build a map of the environment and perform the localization task with a Simultaneous Localization And Mapping (SLAM) approach.

In this work, we consider the problem of exploring three-dimensional (3D) cavities with a tethered aerial robot. Such spaces are often accessed by a narrow entrance tunnel and can be large-scale, posing challenges for autonomy, computation and communication. The latter is important to allow an operator to interact with the robot in real time, e.g. to inspect the wall of an underground mining stope. Tethered Unmanned Aerial Vehicles (UAVs) offer the advantage of unlimited flight time and no loss of communication.

Manuscript received: February, 22, 2022; Revised May, 25, 2022; Accepted July, 6, 2022.

This paper was recommended for publication by Editor P. Pounds upon evaluation of the Associate Editor and Reviewers' comments. This work was supported by the University of Sherbrooke.

[1]The authors are with the Createk Design Lab, University of Sherbrooke, Sherbrooke, J1K 2R1, Canada louis.petit@usherbrooke.ca; alexis.lussier.desbiens@usherbrooke.ca

It is desirable to have a rewindable tether to allow for proper 3D modeling and to avoid getting caught on the ground with a dangling wire. The tether is assumed to be always in taut condition. The extra wire should be stored in an onboard spool, allowing the UAV to continue its mission even if the tether gets caught in obstacles. However, its maximum length should be kept small since it directly affects the drone's payload at takeoff. Our testing platform is the NetherDrone (see Fig. 8a) which fulfills these characteristics except for the rewind which will be integrated in the future.

The approach presented in this paper aims at exploring 3D cavities by minimizing the global path cost and respecting a maximum length of available tether, thanks to a new hierarchical framework composed of a global level of exploration path planning that solves a Traveling Salesman Problem (TSP) and a local level of path planning whose parameters make it possible to adjust the trade-off between the tether unwinding and the path cost. On average, our method of tether-aware path planning for exploration (TAPE) generates a 4.1% increase in distance traveled compared to the TSP solution, with which the length of the tether remains below the maximum allowed value in 53% of the simulated cases against 100% with TAPE.

This paper is organized as follows. Section II defines the problem addressed. In section III, related work regarding autonomous exploration and navigation is described and the main differences are discussed. In Section IV, the proposed approach is presented and detailed. Section V provides results from simulations in various environments, along with a real-world flight with a mapping UAV. A comparative study of the methods is also presented. Finally, this paper concludes with a discussion and potential perspectives.

## II. PRELIMINARY MATERIAL

### A. Assumptions

Any rigid 3D object pose can be defined by a 6-parameter configuration: $(x, y, z, \phi, \theta, \psi)$ and the group of motions is $SE(3) = \mathbb{R}^3 \times SO(3)$. An aerial robot used for environment mapping (e.g. a quadcopter) is likely to move at low speed, thus keeping the roll and pitch parameters ($\phi$, $\theta$) close to zero. Also, the yaw parameter ($\psi$) can be controlled independently. A 3D mapping robot should be equipped with an omnidirectional sensor (i.e. spherical field of view). If not, $\psi$ should be controlled to let the mapping sensors point towards the next exploration goal at all times. The problem can thus be considered holonomic with negligible dynamic constraints. Therefore, for a fixed speed, the mission time is proportional to the distance traveled. Furthermore, after applying a relevant off-line expansion to the occupied space to produce the collision-free space, the problem can be solved as a point-like problem.

Sensor footprints, including LiDARs, typically stop at surfaces unlike e.g. medical ultrasound. Narrow pockets, whose exploration would result in a collision, can therefore not be covered. This results in a residual volume that is not connected to the simply connected set of collision free configurations.

We focus on exploration of cavities, such as mining stopes—whose volume calculation is crucial to optimize stockpile, excavation, and extension operations. They are likely to be 3-ball homotopic by the nature of their creation (i.e. by explosion). The entrance tunnel can be artificially closed to give a bounded volume (see Fig. 2). Environments that are not 3-ball homotopic will also be considered since we cannot guarantee the form of any unknown environment. Environments such as underground tunnel networks are outside the scope of this study and should be explored with more appropriate approaches such as [4]–[6].

In our case, the tether is anchored to a fixed base station located at takeoff position and is assumed to be taut at any time, as it is retractable. In some cases, entanglements (i.e. tether contact with an obstacle) are unavoidable to allow the exploration of an environment. Since the extra wire is stored on-board and does not restrict the movement of the robot in case of a contact with an obstacle, entanglements are authorized, unlike the approach developed in [7].

### B. Notation

We define the world $\mathcal{W} = \mathbb{R}^3$ and consider a tethered aerial robot $\mathcal{A}$, defined in $\mathcal{W}$ as a point, and $\xi_t$ the configuration of $\mathcal{A}$ at time $t$, with $\xi_t = (x_t, y_t, z_t)$.

Let $V \subset \mathbb{R}^3$ be a bounded volume to be explored. $\mathcal{M}$ is an occupancy map representation of all the observable space in $V$, discretized with cubical voxels $m \in \mathcal{M}$, and is incrementally built from observations of the omnidirectional sensor $\mathcal{S}$ with a maximum effective range $d_{max}$.

The search space $V$ can be decomposed into the free space $V_{free}$ and the occupied space $V_{occ}$. The simply connected set of collision free configurations is denoted by $\Xi \subseteq V_{free}$.

Let $\bar{\mathcal{V}}_m \subseteq \Xi$ be the set of configurations from which a voxel $m$ can be perceived from $\mathcal{S}$, and $V_\xi \subseteq V$ the volume perceived by $\mathcal{S}$ at $\xi$.

The residual map $\mathcal{M}_{res} \subseteq \mathcal{M}_{unknown}$ has a volume $V_{res} = \bigcup_{m \in \mathcal{M}} (m \mid \bar{\mathcal{V}}_m = \emptyset)$. Considering $\mathcal{L} \subset \mathbb{R}^3$, the set of configurations along the robot past trajectory, the tether length will be denoted as $L(\mathcal{L})$ and the perceived volume at time $t$ is $V_t = \bigcup_{\xi \in \mathcal{L}_t} V_\xi$.

The goal of the exploration is to identify the free and occupied parts of $V$. The explored volume can then be expressed as $V^E = V^E_{free} \bigcup V^E_{occ} = \bigcup_{\xi \in \mathcal{L}} V_\xi$.

### C. Problem statement

Given an initially unknown bounded volume $V$, incrementally identified in a mapping representation $\mathcal{M}$, find a collision free path $\mathcal{T} : [0, 1] \to \Xi$ that passes through $\xi_1, \xi_2, \ldots$, such that

a) the boundary conditions are respected:

$$\begin{aligned} \mathcal{T}(0) &= \xi_{current}, \\ \mathcal{T}(1) &= \xi_{home}, \end{aligned} \tag{1}$$

b) the tether remains below a maximal length $L_{MAX}$:

$$\max(L(\mathcal{L} \to \mathcal{L} + \mathcal{T}))) < L_{MAX}, \tag{2}$$

c) the observable volume is fully explored:

$$\bigcup_{\xi \in (\mathcal{T} \bigcup \mathcal{L})} V_\xi = V \setminus V_{res}, \tag{3}$$

d) the path cost function $c : \Sigma_\Xi \to \mathbb{R}_{\geq 0}$, which assigns a non-negative cost to the set $\Sigma$ of all paths in $\Xi$ that respect (1)(2)(3), is minimized:

$$c(\mathcal{T}^*) = \min_{\mathcal{T} \in \Sigma_\Xi} c(\mathcal{T}). \tag{4}$$

When the environment is considered fully explored as in (3) or if the tether is insufficient for exploring the whole volume, find a collision-free path $\mathcal{T} : [0, 1] \to \Xi$ that respect (1)(2)(4) to return the robot to its takeoff location.

Due to the a priori unknown nature of the environment, the problem must be solved incrementally as the robot explores and the path must be computed on-line, with the latest sensor readings.

## III. RELATED WORK

According to [8], exploration can be defined as the act of moving through an unknown environment while building a map that can be used for subsequent navigation. This problem has been addressed by several approaches, the most effective of which can be grouped into 4 categories: topological exploration, information theory, random sampling-based methods, and frontier-based methods.

Topological exploration is based on the topological representation of the environment (e.g. with a connectivity graph based on map segmentation) [9]–[11]. With edge and intersection detection, it can be efficient in subterranean tunnel networks or offices by focusing on topological completeness. However, it is not suitable for large cavities, which often cannot be divided into separate sections, and which require map completeness (e.g. for volume calculation).

In information theory, the goal is to move the robot to maximize the amount of new information to expect for the next sensors readings [12]–[14]. These methods are often greedy and global optimality is affected by their myopia.

Random sampling-based methods are based on growing a Rapidly-exploring Random Tree (RRT) [15] or use a similar planner [16]. These methods include Next-Best-View Planner (NBVP) [17] and Graph-Based exploration Planner (GBP) [4]. They are very effective in tunnel-like environments but often suffer from a) randomness and therefore neglect of partially explored areas, or b) greedy behavior when a reward gain is assigned, in the same manner as information theory.

Frontier-based methods rely on the idea to move to a boundary between $V^E_{free}$ and $V_{unknown}$ [8]. These popular exploration methods have been applied to many problems [18], [19]. [8] states that the main question in robotics exploration should be "Given what you know about the world, where should you move to gain as much new information as possible?". This method selects the nearest frontier to allow the robot to incrementally expand its map $\mathcal{M}$. Then, other ways of selecting the appropriate frontier were evaluated (e.g. the largest one, a viewpoint that maximizes the number of observable frontiers) [20]. This approach however does not fully apply on this work since we focus on an optimal exploration mission where we want all the information at the end but the intermediate information is not essential, except perhaps to optimize the rest of the mission. With these frontier selection methods, the problem is not solved in a globally optimal way in the sense of (4). As stated in [20], the appropriate exploration algorithm depends very much on the problem. In this case, one should focus on the completeness of the map and not the speed of discovering the largest part of the map in a greedy way, as many algorithms do.

To optimize the overall path, some frontier-based approaches have considered the selection problem as a variant of the TSP [21], [22]. Other methods such as Technologies for Autonomous Robot Exploration (TARE) also solve this problem for viewpoints [5], [6]. TARE explores in a more complete and rapid way than the methods considered as state of the art such as NBVP [17] and GBP [4]. However, they do not consider the return to home in the optimization of the exploration path once the map is fully explored, in order to satisfy (1). In tunnel-like or office environments, this is of little importance since the entire path is composed of round trips, but in the context of large cavity exploration, it is an important factor to consider.

All methods mentioned so far do not consider the case of a tethered robot, which must respect (2). Research has been done to solve a 2D TSP for given waypoints in a known space with the constraint of a non-entangling path. It is based on the observation of path homotopy classes thanks to their identification by the h-signature [23]. This same identification was used in 2D tangle-free exploration with the nearest frontier method [7]. In 3D, an entanglement may or may not be present for the same homotopy class, so h-signatures are not relevant information [24]. Work has been done to model a tether taut in 3D around an obstacle [25]. This model can be refined by incorporating multi-relaxation, and intermediate relaxation, as discussed in section IV-A.

Most successful exploration models do not consider the case of a tethered aerial robot. Those that do [7] are not optimized in terms of exploration path length and only consider the two-dimensional case. Moreover, the 3D modeling of a tether is close to reality but not reliable in some cases. The goal of this work is to provide a new 3D exploration framework minimizing the mission time in cavities, while respecting a maximum tether length.

## IV. PROPOSED APPROACH

This section starts by giving a method to compute a 3D tether. Then, the steps for processing the frontiers are stated. Finally, the global and local planning strategy are explained.

### A. Tether model

As said in section III, the model defined in [25] can be made more accurate by incorporating multi-relaxation, and intermediate relaxation. Algorithm 1 presents the model after these 2 modifications. The contact points $CP$ must be initialized with the position of the base station. Multi-relaxation is implemented in lines 8 to 11 to deal with the case where several contact points are visible simultaneously for a given waypoint. Also, in 3D, it may happen that the triangle containing (current_contact, last_contact, WP) is not collision free, but the contact point should move in the direction where the robot pulls the tether. Lines 15-18 implement intermediate relaxation via a recursive call to ensure that the line (current_contact, WP) is followed as if it were waypoints, to find the new contact point.

**Algorithm 1:** Tether model

**Input:** Map $\mathcal{M}$, trajectory $\mathcal{L}$, contact points $CP$, max tether length $L_{max}$
**Output:** Final tether length $L$, max tether length $L_{max}$, contact points $CP$

```
1  for every WP in L do
2      current_contact ← CP(n);
3      relax_flag ← false;
4      if CP.size() > 1 then
5          last_contact ← CP(n-1);
6          relax_flag ← not ObstacleConfined(current_contact,
             last_contact, WP);
7      if relax_flag then
8          do
9              CP.pop();
10             repeat lines 2-6;
11         while relax_flag;
12     else
13         if CheckCollision(current_contact, WP) then
14             CP.push(collision_point);
15         else if CP.size() > 1 then
16             RC ← RayCasting(current_contact, WP);
17             if not ObstacleConfined(current_contact,
                 last_contact, RC(1)) then
18                 CP ← PredictTether(RC, CP);
19     L ← CP(n).cost() + dist(CP(n), WP);
20     L_max ← max(L,L_max);
21 return L, L_max, CP;
```

### B. Frontier processing

Similar to the 2D approach in [8], frontiers can be extracted from a 3D occupancy map (OctoMap) [26]. In this map, the voxels are classified in 3 categories: *free*, *occupied* and *unknown*. Frontiers are the voxels in $\mathcal{M}_{free}$ that have at least one neighboring voxel in $\mathcal{M}_{unknown}$.

*Filtering:* The voxels are filtered to satisfy the closed cavity assumption. It can be achieved with a box filter corresponding to the search space $V$. However, to ensure that only the entrance is artificially filtered and for the sake of generality (e.g. for concave environments), a best practice is to use a viewing frustum, as in the FrustumCulling filter of the PCL library. The filter camera is oriented parallel to the ground, and looking backwards with respect to the home pose. The camera position $(x_{cam}, y_{cam}, z_{cam})$ and field of view can be determined at time $t = 0$. For a North-East-Down axis system, the dimensions provided by a vertical slice in $\mathcal{M}_{occ}$ at $x = x_0$ (shown by the red voxels in Fig. 1) allow to fix a) $y_{cam}$ and $z_{cam}$ at the center of this slice, and b) the dimensions of the near plane. Then, the constraint to cover, in the frustum, all the voxels of $\mathcal{M}_{occ}$ located in $x < x_0$, allows to fix $x_{cam}$ as well as the field of view.

*Clustering:* In order to work with a reasonable number of potential views to evaluate, the frontiers are clustered using a Euclidean clustering algorithm in a Kd-tree structure [27]. If the search space is likely to have too many clusters compared to the computational power available for the goal selection procedure detailed in section IV-C, it is preferable to use a k-means clustering algorithm.

*Representation:* Finally, a candidate viewpoint $\xi_{goal,i}$ is generated for the observation of each frontier cluster $C_i$. For the sake of computation time and without loss of performance for large spaces, $\xi_{goal,i}$ is generated as the closest point in $\Xi$ to $m_{c,i}$, which is itself the closest point in $C_i$ to the centroid of $C_i$. Note that the point $m_{c,i}$ is not necessarily equivalent to the centroid of the cluster. For the case of a concave cluster, the true centroid could belong to $V_{occ}$ and/or not allow to observe $C_i$ because of occlusions. The point $\xi_{goal,i}$, as defined above, allows to observe at least partially $C_i$. This is even more true when the clustering is done by euclidean clustering where the voxels are very likely to be neighbors, contrary to a k-mean algorithm where a cluster can be more scattered. In the case where $C_i$ is not entirely visible from $\xi_{goal,i}$, the exploration completeness (3) is ensured by the fact that the voxels not visible in $C_i$ will be taken into account for the next exploration step [22]. For a user whose computation time is not a limitation, a visibility-based approach as in [18] can be used instead, but it is difficult to implement in real time.

### C. Global path planning

With a large enough sensor range, it is possible to have a relatively good a priori information of the space to explore. Moreover, since cavities are simply connected topological spaces, exploring a tunnel or pocket that has not yet been perceived requires crossing one of the known frontiers to access it, and probably ultimately returning to it when the exploration of that area is complete. The distance traveled in that area is therefore often added to the travel distance, regardless of the order in which the frontiers are visited.

Under these assumptions, the solution that optimizes the path length of visiting the frontiers—given what you know at time $t$—is substantially close to the global solution and the problem can be viewed as a TSP. The only difference with a TSP is that the departure city (i.e. the drone's position) is not equivalent to the arrival city (i.e. the home position).

Other than that, the problem can be solved using state of the art algorithms. For a high number of clusters, computing an exact solution ($\approx \mathcal{O}(n^2)$) can be demanding but approximate algorithms give very close results in a reasonable time. The Lin-Kernighan heuristic is used to solve the TSP as it is one of the most powerful algorithms [28]. To increase the fidelity of the solution, the adjacency matrix used to solve the problem is based on the path cost between the different $\xi_{goal,i}$ positions. The cost can be computed with any path planner but RRT-Rope is preferred since it rapidly computes a near-optimal path in large uncluttered spaces [29].

The efficiency of the approaches presented in [4] and [5] comes largely from their hierarchical framework. In 3D, as shown later in the section V-A by trying an approach to solving the TSP to minimize entanglements similar to [23], the visiting order of the candidates in the TSP has little influence on the entanglements or the length of the tether, while the local behavior of the robot has a strong influence on these parameters. The approach presented here is therefore based on a global and a local level. The global planning takes into account the boundary conditions (1), the observation of all frontiers (3) and the minimization of the path (4) as a goal for solving the TSP, while the local planning takes into account the length of the tether (2) and the minimization of the path (4). This allows the method to run easily in real time as the tether is only considered locally. Replanning is performed at each map update to ensure path quality. Algorithm 2 presents the global exploration process. The function *LocalPlanner* is detailed below.

**Algorithm 2:** Exploration path planner

**Input:** map $\mathcal{M}$, collision free space $\Xi^E$, positions $\xi_{home}$, $\xi_{current}$, $\xi_{goal}$
**Output:** exploration path $\mathcal{T}$

1 $F \leftarrow$ ExtractFrontiers($\mathcal{M}$);
2 $F$.Filter();
3 **if** $F$.size() == 0 **then**
4 | $\mathcal{T} \leftarrow$ LocalPlanner($\xi_{current}$, $\xi_{home}$, $rth$=true);
5 **else**
6 | $C_{1,\dots,n} \leftarrow$ Cluster($F$);
7 | $\xi_{goal,\,1,\dots,n} \leftarrow$ Represent($C_{1,\dots,n}$);
8 | $\xi_{goal} \leftarrow$ SolveTSP($\xi_{goal,\,1,\dots,n}$, $\xi_{current}$, $\xi_{home}$);
9 | $\mathcal{T} \leftarrow$ LocalPlanner($\xi_{current}$, $\xi_{goal}$);
10 **return** $\mathcal{T}$;

### D. Local path planning

Once the TSP is solved, a goal candidate $i$ is chosen and a path query is sent to the local planner to go to $\xi_{goal,i}$. The local path is based on the generation of 5 possible trajectories ($\mathcal{T}_G$, $\mathcal{T}_{HG}$, $\mathcal{T}_{CPG}$, $\mathcal{T}^t_{HG}$, $\mathcal{T}^t_{CPG}$) whose choice is determined by a decision function. These trajectories result from 2 steps: a) searching for path portions, and b) assembling and shortcutting. The path portions queries between 2 points can be done with any path planner but RRT-Rope [29] is preferred here. The procedure is detailed below.

*Path options:* $\mathcal{T}_G$ is the result of the path request from the drone's position $\xi_{current}$ to the next goal candidate $\xi_{goal}$. $\mathcal{T}_{HG}$ is obtained by searching 2 portions of path $\tau_1(\xi_{current}, \xi_{home})$, and $\tau_2(\xi_{home}, \xi_{goal})$. These portions are assembled and a deterministic shortcut as in [29] is applied on the resulting path. $\mathcal{T}^t_{HG}$ is obtained in the same way but the shortcut applied is different. It is the same algorithm as for the calculation of the tether for given waypoints. Thus, the path obtained remains stuck in the local minima where the tether may get stuck. $\mathcal{T}_{CPG}$ and $\mathcal{T}^t_{CPG}$ are similar to $\mathcal{T}_{HG}$ and $\mathcal{T}^t_{HG}$ but take as input portions of path that pass through the contact points in the reverse order, i.e. $\tau_0(\xi_{current}, CP_n)$, $\tau_1(CP_n, CP_{n-1})$, ..., $\tau_n(CP_1, \xi_{home})$, $\tau_{n+1}(\xi_{home}, \xi_{goal})$.

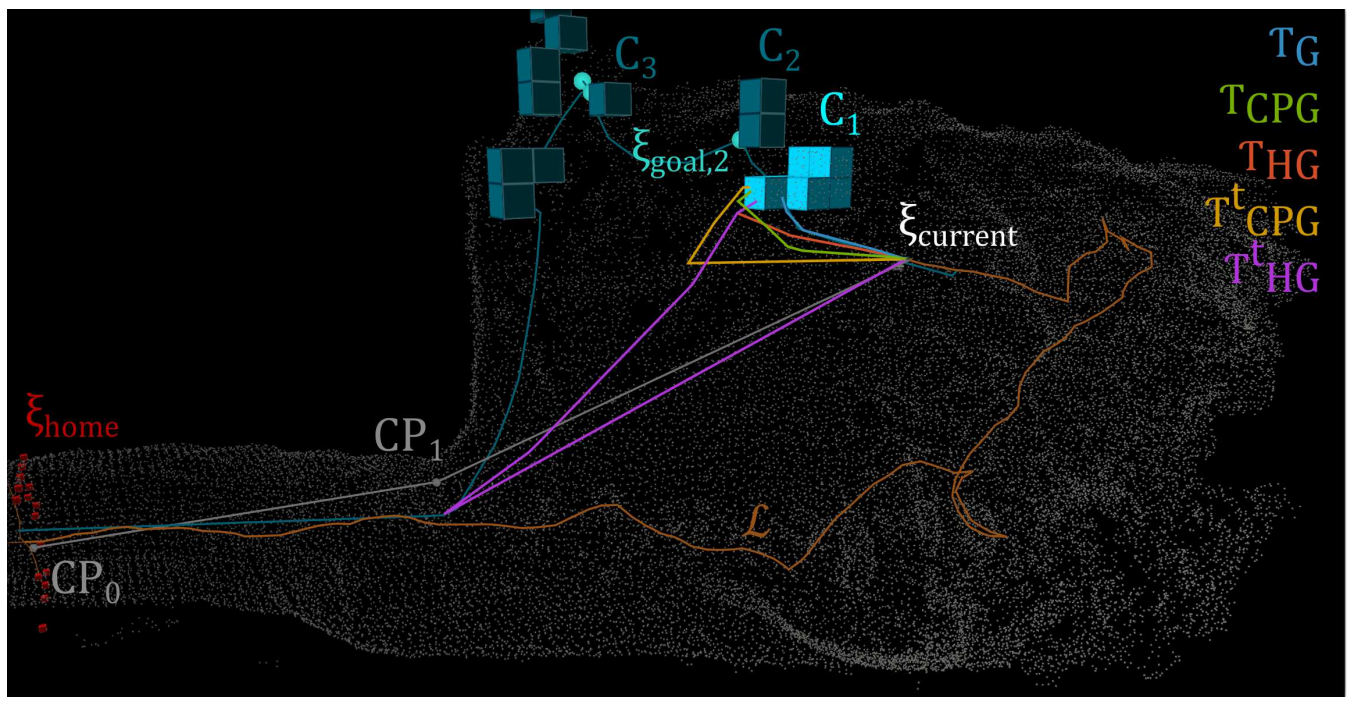

Fig. 1: An example of exploration in Env. 1. The path options are in blue, red, green, yellow and purple. The past trajectory is in brown, the contact points and tether model in grey. The global exploration path and the frontiers clusters are in dark blue with the viewpoints and the next cluster chosen in cyan.

For all paths, shortcuts are applied after assembly, and the assembly points (i.e. home or contact points) are not retained in the final path. The 5 possible paths are shown in Fig. 1. These paths lead to an informed range of options from the shortest ($\mathcal{T}_G$) to the one that allows the total untangling of the tether ($\mathcal{T}^t_{CPG}$), through options that allow to partially untangle the tether while remaining relatively short. If there is only one contact point in the current tether model (i.e. $CP = \{\xi_{home}\}$), $\mathcal{T}_{CPG}$ and $\mathcal{T}^t_{CPG}$ will not be evaluated. It is also trivial that for the case of a return to home, $\xi_{goal} = \xi_{home}$. Then, $\mathcal{T}_{HG}$ and $\mathcal{T}^t_{HG}$ will not be calculated.

*Decision function:* The decision function for a past trajectory $\mathcal{L}$ and a planned path $\mathcal{T}$ is defined as

$$\begin{aligned} f(\mathcal{L}, \mathcal{T}) &= k_1 \times c(\mathcal{T}) \\ &+ k_2 \times max(L(\mathcal{L} \to \mathcal{L} + \mathcal{T})) \\ &+ k_3 \times L(\mathcal{L} + \mathcal{T}) \\ &+ k_4 \times CP(\mathcal{L} + \mathcal{T}), \end{aligned} \tag{5}$$

whose terms are respectively the path cost $c(\mathcal{T})$ (i.e. distance to cover), the maximum tether length between $\xi_{current}$ and $\xi_{goal}$, the tether length at $\xi_{goal}$ and the number of contact points at $\xi_{goal}$. In practice, the computation of the function $L$ and the associated $CP$ is only performed for $\mathcal{T}$ since the $CP$ associated to $\mathcal{L}$ are kept in memory. These parameters allow us to make an informed decision that takes into account the distance ($k_1$) and the maximum tether length ($k_2$), and also to locally relax the problem ($k_3$, $k_4$) to stay far away from $L_{MAX}$ for future path queries. A sensitivity analysis for these parameters is performed in section V-A.

Algorithm 3 presents the approach used to compute a local path. Note that the function *RopeShortcut* uses only the shortcut described in [29] with a custom path to replace lines 1 through 3 of the RRT-Rope algorithm. Also, the function *CostFunction* is based on (5) after evaluating the tether-related variables as in section IV-A. The function *TetherShortcut* uses the same tether model but for a planned path instead of a past trajectory, and the collision tests are performed in the collision free space $\Xi^E$ instead of $\mathcal{M}_{free}$ in order to generate a valid trajectory.

**Algorithm 3:** Local path planner

**Input:** map $\mathcal{M}$, collision free space $\Xi^E$, positions $\xi_{home}$, $\xi_{current}$, $\xi_{goal}$, contact points $CP$, RTH flag $rth$
**Parameters:** $k_1$, $k_2$, $k_3$, $k_4$
**Output:** local path $\mathcal{T}$

1 $\mathcal{T}_G \leftarrow$ RRT-Rope($\xi_{current}$, $\xi_{goal}$);
2 **if** $CP$.size() $= 1$ **and** $rth$ **then**
3     **return** $\mathcal{T}_G$;
4 $\tau_2 \leftarrow$ RRT-Rope($\xi_{home}$, $\xi_{goal}$);
5 **if not** $rth$ **then**
6     $\tau_1 \leftarrow$ RRT-Rope($\xi_{current}$, $\xi_{home}$);
7     $\mathcal{T}_{HG} \leftarrow$ RopeShortcut($\tau_1$ + $\tau_2$);
8     $\mathcal{T}^t_{HG} \leftarrow$ TetherShortcut($\tau_1$ + $\tau_2$);
9 **if** $CP$.size() $> 1$ **then**
10     $\tau_1 \leftarrow$ RRT-Rope($\xi_{current}$, $CP_n$);
11     **for** $i \leftarrow 0$ **to** $CP$.size() **do**
12         $\tau_1$.add(RRT-Rope($CP_{n-i}$, $CP_{n-i-1}$));
13     $\mathcal{T}_{CPG} \leftarrow$ RopeShortcut($\tau_1$ + $\tau_2$);
14     $\mathcal{T}^t_{CPG} \leftarrow$ TetherShortcut($\tau_1$ + $\tau_2$);
15 $\mathcal{T}_{options} \leftarrow$ ($\mathcal{T}_G$, $\mathcal{T}_{HG}$, $\mathcal{T}^t_{HG}$, $\mathcal{T}_{CPG}$, $\mathcal{T}^t_{CPG}$);
16 **return** $\mathcal{T}_{options}$(MinElement(CostFunction($\mathcal{T}_{options}$)));

## V. RESULTS

### A. Simulation studies

The approach presented in this paper was deployed on ROS (Robot Operating System) and tested on 4 different simulated environments that were reconstructed from real LiDAR scans. The simulated UAV shown in Fig. 8a is equipped with a rotating Ouster LiDAR and has a quasi-omnidirectional field of view after a full rotation. A SLAM approach called RTAB-Map [30] is used to provide localization data and the OctoMap from which the frontiers can be extracted. Simulations were performed with an HP Z440 Workstation with 12 Intel Xeon processors running at 3.5 GHz, 24 GB of RAM, with a node frequency of 10 Hz for the global planner and 15 Hz for local path planning.

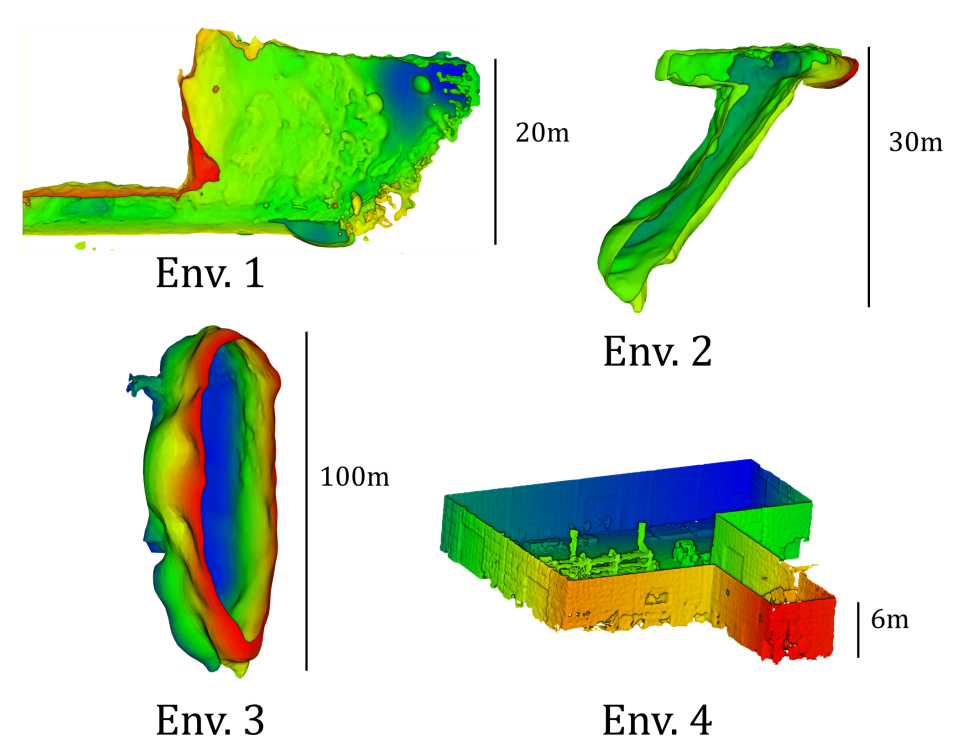

Fig. 2: Cross-section view of the simulation environments.

The 4 environments pictured in Fig. 2 have been chosen to best represent the spatial characteristics of some typical exploration missions. *Env. 1* and *Env. 2* represent classic underground mining stopes with an access tunnel and narrow passages to navigate in order to map certain pockets. *Env. 2* is a bit narrower while *Env. 1* extends on an inclined plane at the end of the tunnel. *Env. 3* is also an underground mining stope with an access tunnel but is much larger, with some pockets in the walls. *Env. 4* is a cluttered indoor environment with shelves, pipes and other obstacles, to challenge the performance of the algorithm in environments with high risk of tether entanglements.

*Performance comparison:* Fig. 3 shows the results of the comparison of different algorithms: the selection of the biggest frontier in blue, the nearest frontier in red, the nearest frontier with our local planner in yellow, the resolution of a TSP in purple, our algorithm in green, and the resolution of a TSP taking into account the tether similar to [23] without our local planner in cyan and with our local planner in burgundy. The first two algorithms are mainly used as an order of magnitude since it has already been proven that the TSP produces a shorter path [21]. The other state of the art algorithms are not compared since such studies already exist [20], and there is none to our knowledge that takes into account the tether for a 3D path.

In all environments, the algorithms that use our local planner do not exceed $L_{MAX}$ in 100% of the cases on 32 samples and the objective (2) is therefore achieved.

As announced in section IV-C, *gTAPE* results in tether lengths similar to the other methods because the global path has little influence on entanglements. Also, it often gives a longer path because a) it sometimes leads to local hesitations between several successive iterations and b) it only takes into account the use of one planner to move between the frontiers because it would be too computationally demanding to consider the 5 path options for the $N \times (N-1)/2$ branches of the graph. With our local planner, *gTAPE** also suffers from this global distance increase.

In *Env. 1* and *Env. 3*, the 2 algorithms using the TSP (i.e. *TSP* and *TAPE*) are the shortest. In *Env. 2*, the path lengths are very similar between the TSP algorithms and the greedy algorithms, since the environment is quite narrow and the paths are likely to be similar. In *Env. 3* and *Env. 4*, we notice that our local planner leads to a slight increase in distance traveled when comparing *nearest* and *TSP* with *nearest** and *TAPE* respectively.

Fig. 4 shows 3 paths taken at random from the same dataset, for the greedy algorithm, the TSP and our algorithm. In Fig. 3g, we notice that the increase in distance avoids entanglements in the shelves located in the middle of the environment, as can be seen in Fig. 4b by observing the obstacle bypassing by our algorithm compared to the TSP. This behavior is also observed in Fig. 4a where our planner allows to go around the rocky outgrowth located on the top right. On average, our method generates a 4.1% increase in distance traveled compared to the globally optimal method (TSP), which entangles the tether in 46.9% of cases. In Fig. 3g, we observe the limits of the TSP for global path optimization, as the possibility of discovering a frontier from behind after bypassing a column is not well treated. It is in the limit of the case of a 3-ball homotopic environment, as explained earlier in the section II-A. In *Env. 1*, *Env. 2* and *Env. 3*, a comparison of the distance traveled show the usefulness of using the TSP over the greedy algorithm to optimize the global path.

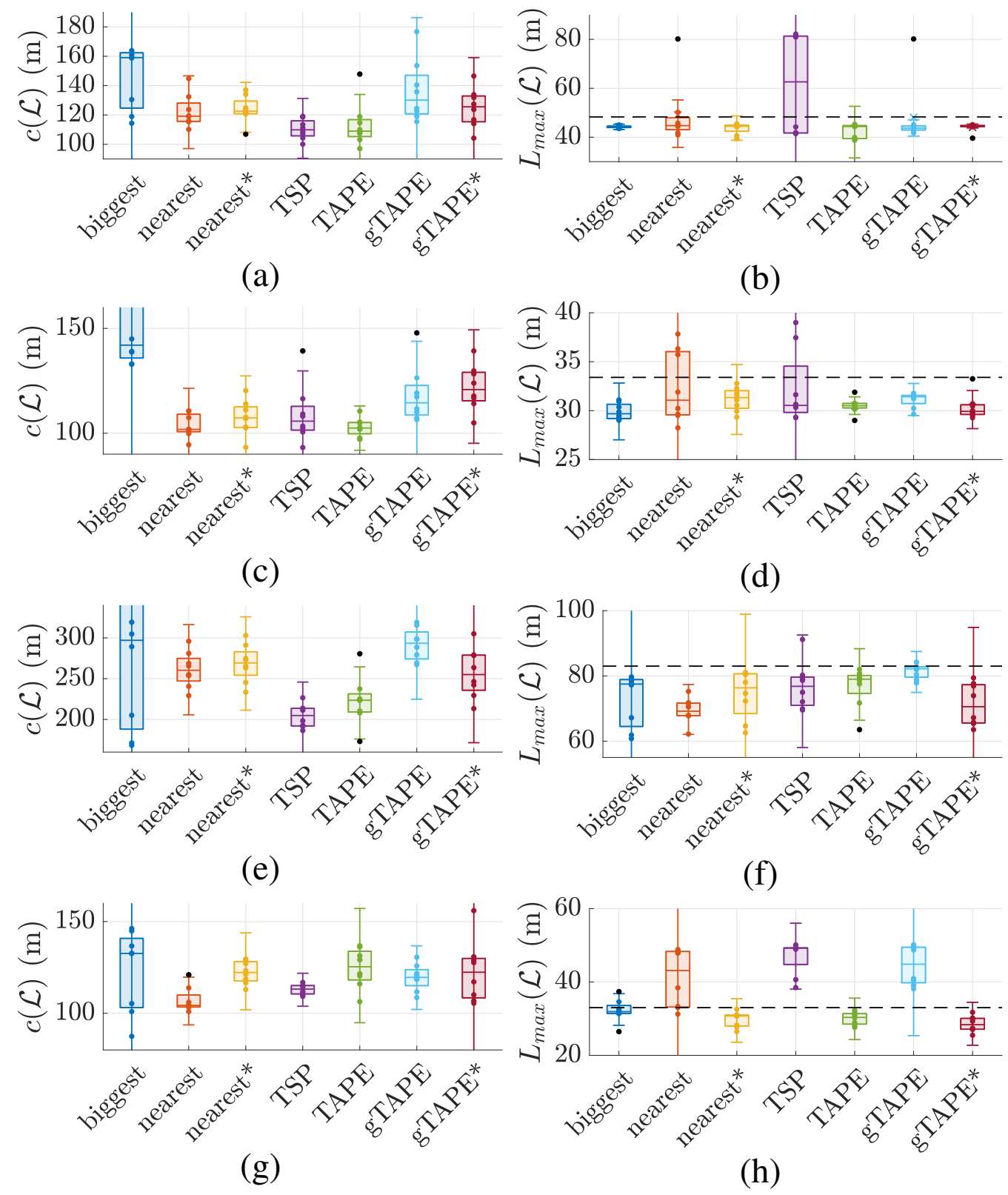

Fig. 3: Comparison of covered distance and maximum tether length for different algorithms in a-b) Env. 1, c-d) Env. 2, e-f) Env. 3, g-h) Env. 4. The dotted line shows the maximum tether length $L_{MAX}$ needed to travel the longest distance diagonally to the end of the cavity from the base station, taking the shortest path. The data are presented as a boxplot for 8 samples in each category, with the cross representing the average, and the black dots representing the outliers. The parameters $k_i$ used in (5) are (1, 7, 1, 10).

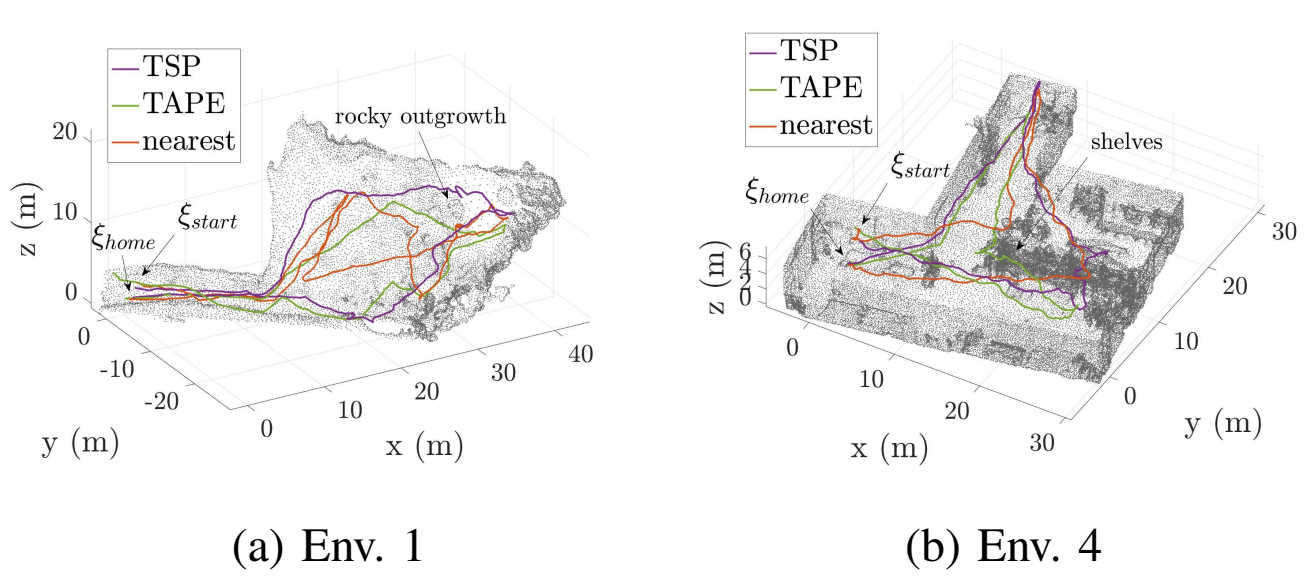

(a) Env. 1 (b) Env. 4

Fig. 4: Exploration path display.

Finally, a feature that was not stated in the initial objectives is the possibility to recover the tether after the mission. In the case of an unsafe or inaccessible environment for humans, a stuck tether results in a loss of material. Fig. 5 shows the final length of the tether at the end of the exploration mission for the same samples as in Fig. 3. Our method recovers the tether completely in 96.9% of the cases and results in a loss of only $10m$ in the only outlier case among the 32 trials.

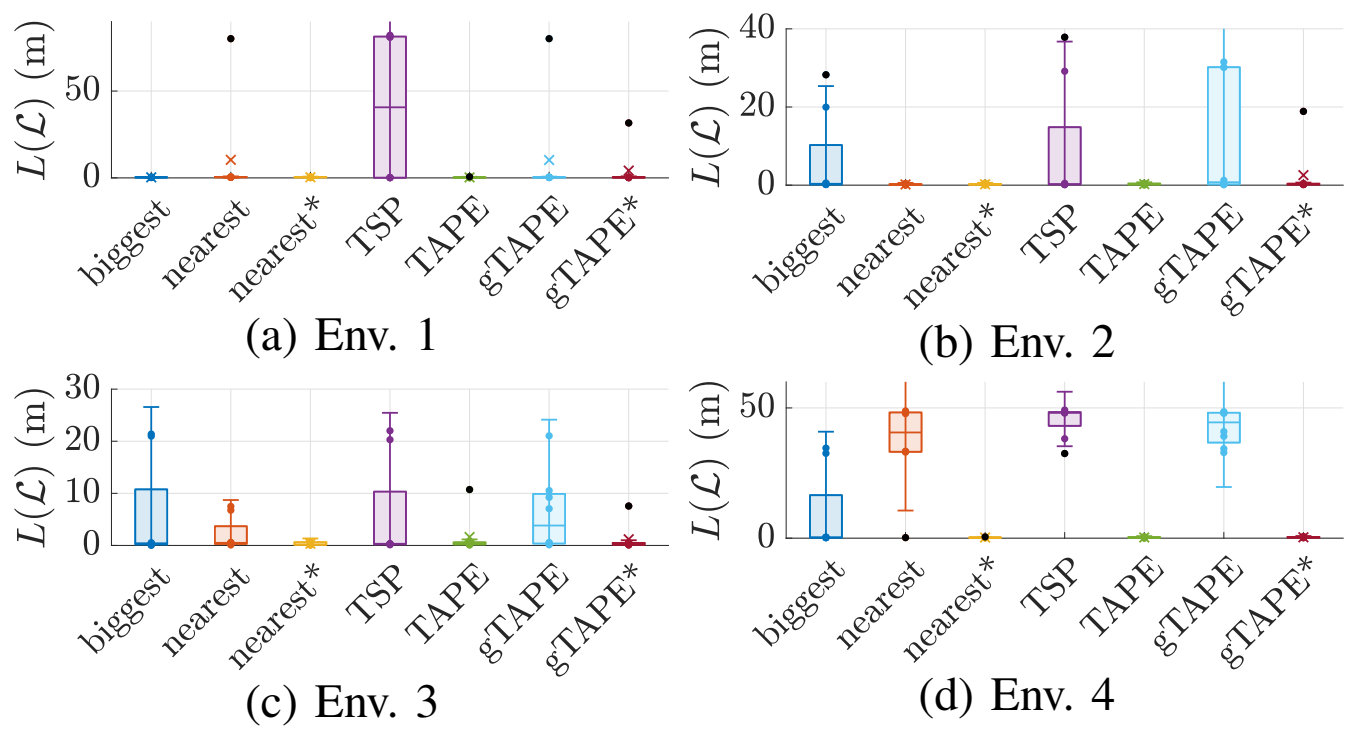

(a) Env. 1 (b) Env. 2 (c) Env. 3 (d) Env. 4

Fig. 5: Tether length $L$ at the end of the exploration mission.

*Parameters sensitivity:* Fig. 6 and Fig. 7 presents the results of simulations performed with different sets of parameters $k_{1\to4}$ in (5), with the aim of confirming their importance and giving the user a preliminary idea of how to choose them. The tests are performed in *Env. 2* as it is the smallest, with 20 samples per parameter set. This is where we will see a greater sensitivity in the parameters because the distances to be traveled and the tether lengths are shorter and one term of (5) can more quickly become important compared to another.

Fig. 6 shows the effects of variations of $k_1$. Fig. 6a shows that for a large $k_1$, $\mathcal{T}_G$ is chosen more often and the paths with a tether-shortcut less and less. The effect is a decrease of the path cost (Fig. 6b), and an increase of the final and maximum tether length (Fig. 6c and Fig. 6d). In our case, $k_1 = 1$ allows to decrease the path cost without a consequent increase of the tether length.

Applying the same approach, we see that $k_{2\to4}$ each increase the path cost (Fig. 7a). $k_2$ significantly affects $c$ only for $k_2 \leq 10$, decreases $L$ and $L_{max}$ by preventing entanglements

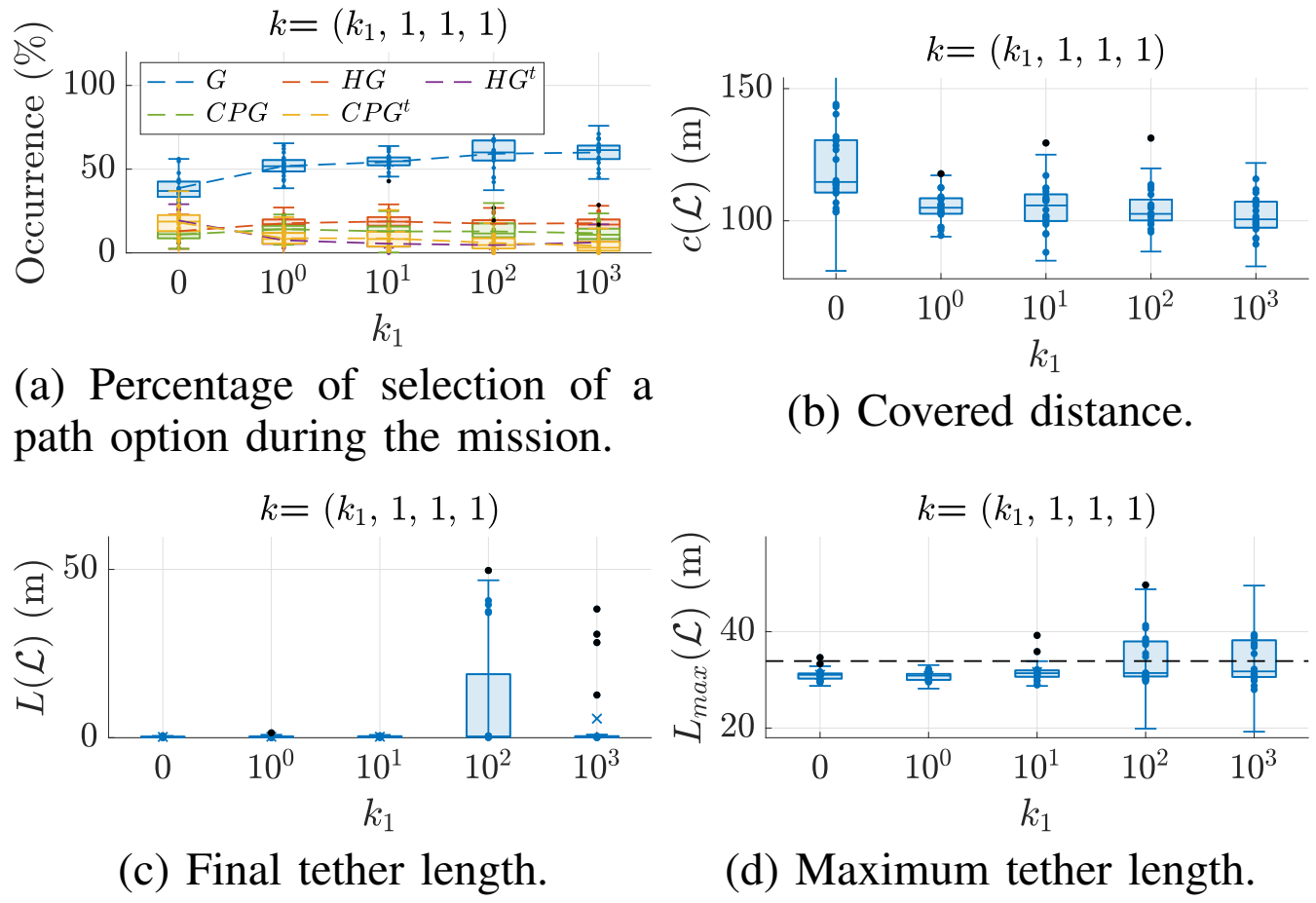

(a) Percentage of selection of a path option during the mission. (b) Covered distance. (c) Final tether length. (d) Maximum tether length.

Fig. 6: Results of the parametric analysis in Env. 2, for different values of $k_1$.

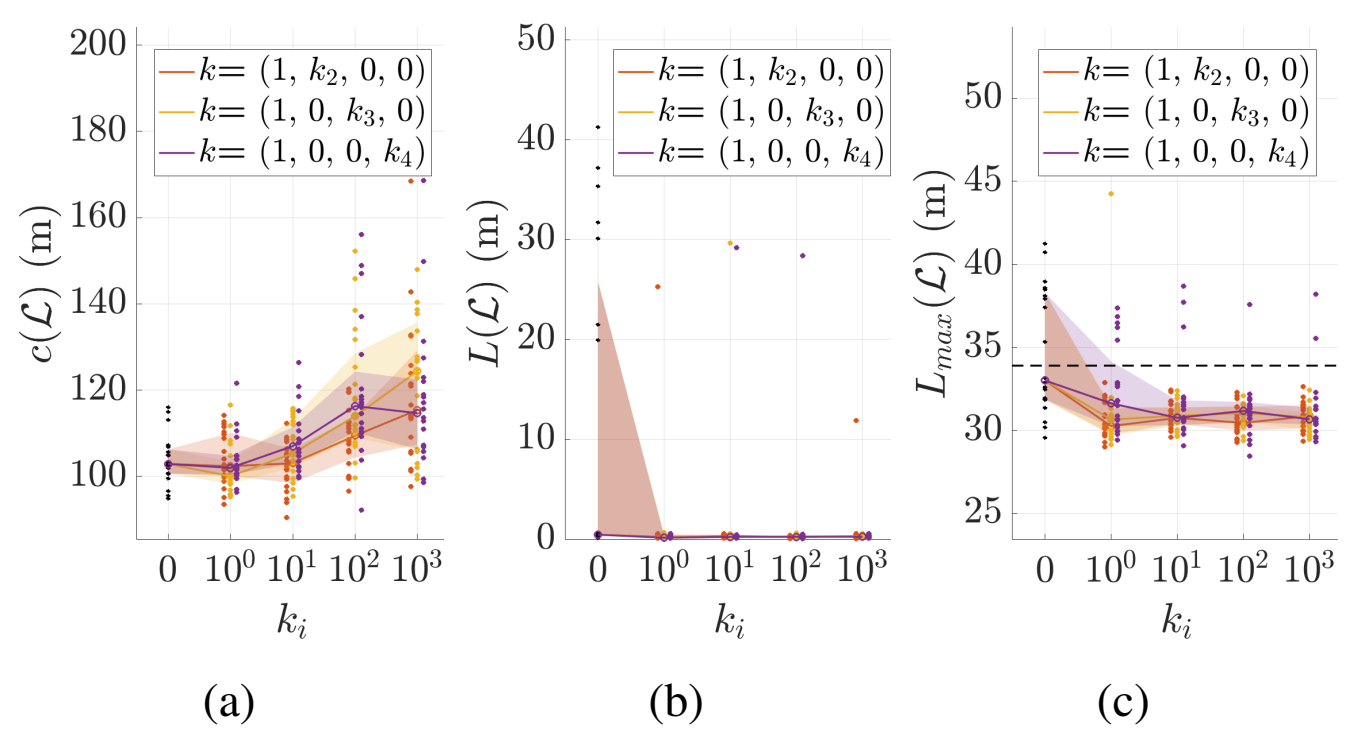

(a) (b) (c)

Fig. 7: Results of the parametric analysis in Env. 2, for different values of $k_2$, $k_3$, $k_4$: a) covered distance, b) final tether length, c) maximum tether length.

when $k_2 \geq 1$. $k_3$ affects $c$ a lot, decreases $L_{max}$, and $L \simeq 0$ for $k_3 \geq 10$. $k_4$ affects significantly $c$, decreases $L_{max}$, and $L \simeq 0$ when $k_4 \geq 100$. A set around $k = (1, 10, 1, 10)$ is thus a good choice for our case. A high precision is not necessary since a small variation of the parameters does not influence the results very much.

### B. Experimental evaluation

Finally, our method is used by the NetherDrone (see Fig. 8a) to explore an unknown indoor environment. Fig. 8b shows the result of this flight. The UAV explores a volume of $6000m^3$ in $164s$, covering a distance of $74m$ with a maximum tether length of $39m$. For operation safety reasons, a part of the environment on the left was marked as a no-go zone with a box filter. The results show that the drone follows a path that avoids the tether getting stuck between obstacles, while keeping this path fairly short.

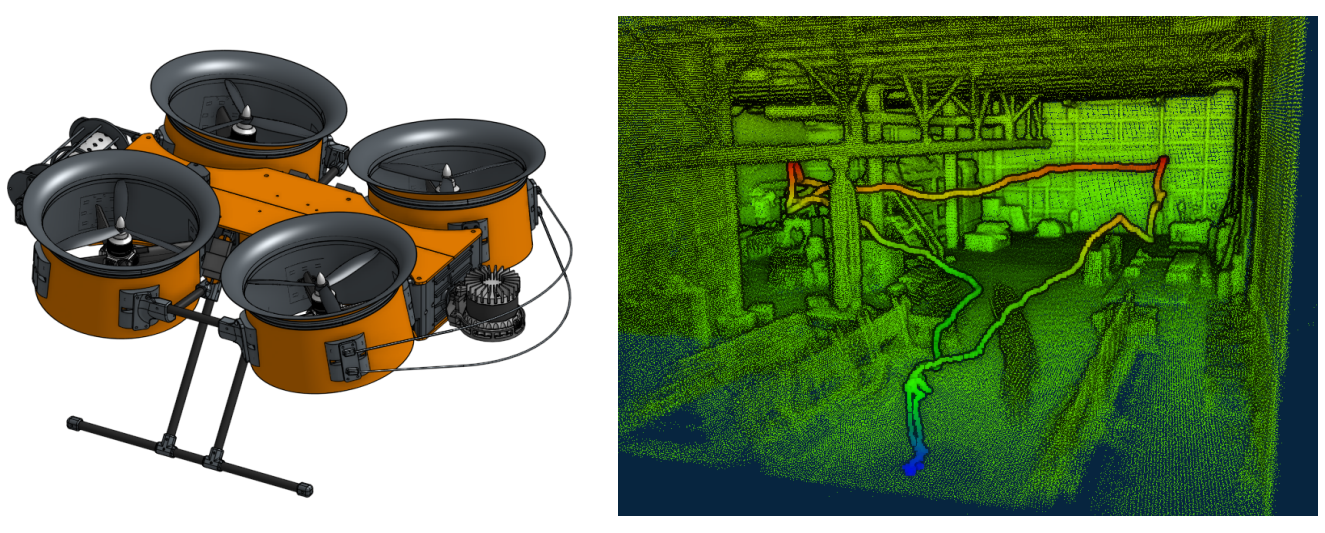

(a) NetherDrone 3D render. (b) Exploration path display.

Fig. 8: Real-world flight experiment.

## VI. DISCUSSION

A remaining limitation of the tether model is that collisions are only considered with obstacles detected by the sensor and not with the prediction of the tether itself, but this rare case did not occur among all our observations, thanks in part to the fact that entanglements are minimal with our approach.

A UAV that does not respect the taut tether assumption can do well with our algorithm if the voxel resolution is adapted to the environment complexity. In our test with the NetherDrone, we observed on the ground that the unrolled tether on the return path was always located within at most $1m$ of the unrolled tether on the outbound path, and it can then be easily rewound. This may reach its limits in complicated environments where the computation will become too expensive, or the mapping inaccuracies too noticeable.

Also, our method performs very well in cavities but might reach its limits for a large maze environment, where another algorithm than RRT-Rope [29] can be used for path generation but the integration in real time becomes challenging.

Regarding adaptability, it has been shown that the global path has little influence on the tether entanglements but is important to minimize the path distance, and the local planner can thus be integrated with other approaches such as GBP or TARE, instead of the frontier-TSP.

## VII. CONCLUSION

In this work, a hierarchical framework for path planning for 3D cavity exploration by a tethered aerial robot is proposed. The global planning focuses on minimizing the distance traveled by solving the Traveling Salesman Problem for clusters of frontiers. The local planning aims at minimizing the distance while avoiding tether entanglements, according to a decision function whose parameters allow to adjust the tether length and path cost trade-off. Our method is evaluated in simulation and in a physical environment against other exploration methods, using the NetherDrone equipped with a rotating LiDAR. Our two-stage architecture is the first to our knowledge to focus on 3D exploration with a tether. The results confirm that it performs better in satisfying path minimization while respecting an acceptable deployed tether length. On average, our method generates a 4.1% increase in distance traveled compared to the method without our local planner, with which the length of the tether remains below the maximum allowed value in 53% of the simulated cases against 100% with our method. Future work plans to deploy this algorithm in more real-world environments.